\documentclass[letterpaper, 10 pt, conference]{ieeeconf}  

\IEEEoverridecommandlockouts                              

\usepackage{amsmath} 
\usepackage{amssymb}  
\usepackage{hyperref} 
\usepackage{graphicx} 
\usepackage{xcolor} 
\usepackage{booktabs} 
\usepackage{algorithm} 
\usepackage{algpseudocode} 
\usepackage{siunitx} 
\usepackage{multirow} 

\title{\LARGE \bf
3D-CovDiffusion: \\ 3D-Aware Diffusion Policy for Coverage Path Planning
}

\author{
Chenyuan Chen$^{1}$, Haoran Ding$^{1}$, Ran Ding$^{1}$, Tianyu Liu$^{1}$, Zewen He$^{1}$, 
\\ Anqing Duan$^{1}$, and Yoshihiko Nakamura$^{1}$%
\thanks{\scriptsize $^{1}$Authors are at Mohamed bin Zayed University of Artificial Intelligence (MBZUAI), Abu Dhabi, UAE.
Email: {\ttfamily \{firstname.lastname\}@mbzuai.ac.ae}
\noindent\textbf{Project page: }
\nolinkurl{https://crystalccy1.github.io/3D-CovDiffusion/}}
}

\hypersetup{
    pdfauthor={Chenyuan Chen, Haoran Ding, Ran Ding, Tianyu Liu, Zewen He, Anqing Duan, Yoshihiko Nakamura},
    pdftitle={3D-CovDiffusion: 3D-Aware Diffusion Policy for Coverage Path Planning},
    hidelinks
}

\begin{document}

\maketitle
\thispagestyle{empty}
\pagestyle{empty}

\begin{abstract}

Diffusion models have shown strong potential for robot skill learning, yet their role in coverage path planning remains underexplored. In industrial surface processing (painting, polishing, spray coating), high coverage requires globally ordered, temporally coherent trajectories rather than stitching unordered local segments. We reformulate coverage path planning as conditional sequence generation and adopt a geometry-conditioned diffusion framework that synthesizes continuous trajectories directly from raw 3D point clouds. Our method produces temporally ordered trajectory chunks and avoids post-hoc heuristic ordering or stitching in prior learning-based methods via simple sequential concatenation, improving sequence-level consistency. A single shared policy generalizes across different geometries without category-specific architectures. Extensive benchmarks show substantial gains over prior learning-based baselines: 98.2\% lower point-wise Chamfer Distance (lower is better), 97.0\% lower jerk (smoother trajectories), and +67.5 percentage points overlapping surface coverage on average.

\end{abstract}

\section{INTRODUCTION}

Industrial surface-processing tasks such as robotic painting, polishing, and spray coating require robots to generate continuous, temporally coherent trajectories that achieve high surface coverage over complex geometries~\cite{nieto2023autonomous, trigatti2018new}. In these applications, task success is not determined solely by local motion accuracy, but by how effectively the robot performs globally consistent surface traversal over time. In practice, high-quality coverage critically depends on generating temporally ordered trajectories rather than locally accurate but unordered motions. Imitation learning~\cite{osa2018algorithmic} has emerged as a promising paradigm for acquiring such skills from expert demonstrations~\cite{tiboni2023paintnet, 3d_diffusion_policy, brohan2022rt, chi2023diffusion, ho2016generative}. However, despite recent progress, existing learning-based approaches to industrial spray painting still face limitations: many methods decompose continuous trajectories into short, fixed-length segments that are predicted independently and later combined using post-hoc heuristic ordering and stitching~\cite{tiboni2025maskplanner, tiboni2023paintnet}. While this simplifies learning, it often disrupts temporal consistency and leads to fragmented trajectories and unstable surface coverage, particularly for complex geometries.

\begin{figure}[t]
    \centering
    \includegraphics[width= 1 \linewidth]{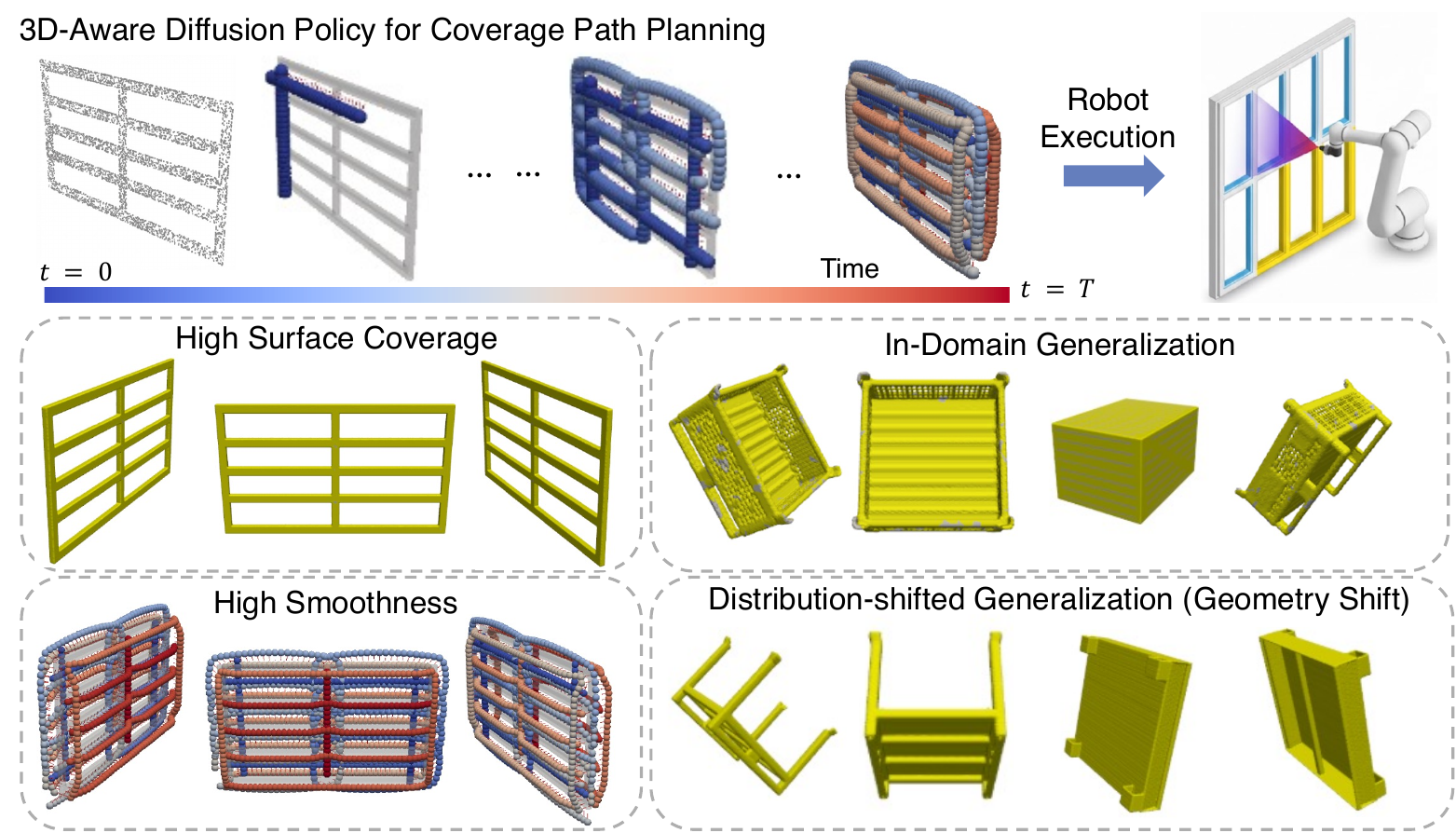}
    \caption{We present 3D-CovDiffusion, a 3D-aware diffusion policy that generates ordered coverage paths, which jointly synthesizes complete trajectories over time ($t=0$ to $t=T$) and yields high surface coverage and smooth trajectories, enabling within-category geometry shift no category-specific fine-tuning.}
    \label{fig:first_figure}
    \vspace{-1.5em} 
\end{figure}

This mismatch arises because surface coverage is fundamentally a sequence-level property that depends on temporal ordering, yet prior learning-based methods often treat ordering as a post-processing step after segment-wise prediction~\cite{tiboni2025maskplanner, tiboni2023paintnet}. To address this challenge, we reformulate coverage path planning as conditional sequence generation and propose a geometry-conditioned diffusion policy that directly synthesizes continuous, temporally ordered trajectories from raw 3D point clouds. Our approach outputs ordered trajectory chunks and avoids post-hoc heuristic ordering/stitching used in prior learning-based methods via simple sequential concatenation. Because diffusion models refine the full trajectory sequence iteratively~\cite{chi2023diffusion}, they naturally support temporal coherence and can yield smoother, more consistent trajectories, making them well suited for sequence-level trajectory generation. Our method conditions trajectory generation on raw 3D point clouds, providing a simple yet expressive representation of object geometry without relying on category-specific priors, as shown in Figure~\ref{fig:first_figure}. As a result, a single diffusion policy generalizes across the benchmark geometries without category-specific architectures. Experimental results on industrial spray-painting benchmarks demonstrate substantial improvements in surface coverage, trajectory smoothness, and geometric fidelity compared to prior learning-based methods.

In summary, by treating coverage path planning as a sequence-level generation problem rather than a post-hoc assembly task, our approach provides a scalable alternative to segment-wise learning pipelines that rely on heuristic ordering and stitching. Our main contributions are summarized as follows:

\begin{itemize}
    \item We cast industrial coverage path planning as conditional sequence generation, modeling trajectory ordering at the sequence level and treating coverage as a trajectory-level property.
    \item We adopt a geometry-conditioned diffusion policy that generates temporally ordered and smooth 6-DoF trajectory chunks, avoiding post-hoc heuristic ordering or stitching in prior learning-based methods via simple sequential concatenation.
    \item We show that a single shared policy generalizes across the benchmark geometries (including within-category geometry shift) and improves coverage, smoothness, and geometric accuracy on the main benchmark over prior learning-based baselines.
\end{itemize}

\section{RELATED WORK}

Coverage path planning for industrial surface-processing tasks has been studied from geometric planning, learning-based trajectory prediction, and more recently, generative policy learning \cite{CPPreview}. In this section, we review prior work by focusing on how different approaches \emph{conceptualize and represent coverage} rather than on specific model architectures.

\paragraph{Geometric Coverage Path Planning}

Traditional coverage path planning methods formulate coverage as an explicit geometric objective or constraint, typically relying on CAD models, surface parameterization, or sampling-based optimization~\cite{CPPreview, chen2002automated, CAD-guided_robot, Atkar-2005-9354, andulkar2015incremental, chen2020trajectory, 9736331OptimizedTrajectories, englot2012sampling}. These approaches can guarantee coverage completeness under ideal modeling assumptions and provide strong interpretability and controllability. However, they often require accurate object models, careful parameter tuning, and extensive manual engineering, which limits their applicability in unstructured or perception-driven settings. Moreover, their reliance on handcrafted rules makes it difficult to integrate them into learning pipelines or to generalize across diverse geometries.

\paragraph{Learning-based Segment and Stroke Prediction for Surface Processing}

Recent learning-based approaches have explored predicting spray painting or surface-processing motions directly from 3D observations~\cite{yuan2018pcn, alliegro2021denoise, ni2020pointnet++} using imitation learning~\cite{tiboni2025maskplanner, tiboni2023paintnet} or reinforcement learning~\cite{kiemel2019paintrl}. These methods typically decompose continuous trajectories into short, fixed-length segments or strokes that are predicted independently and later combined using heuristic ordering or stitching rules. While this segment-wise formulation simplifies learning and allows for flexible local prediction, it treats global temporal ordering as a post-processing step rather than as part of the generation process. As a result, surface coverage becomes an implicit outcome of local predictions instead of a property of the generated trajectory itself, often leading to fragmented paths and unstable coverage. From a representation perspective, these approaches adopt local or unordered output structures that are misaligned with the sequence-level nature of coverage path planning, where coverage quality critically depends on global temporal consistency.

Prior learning-based approaches for industrial surface processing typically predict either unordered surface points or short trajectory segments and then rely on post-hoc heuristics to assemble executable paths.
PaintNet~\cite{tiboni2023paintnet} and MaskPlanner~\cite{tiboni2025maskplanner}, for example, decompose demonstrations into fixed-length segments and apply heuristic ordering/stitching at inference.
In contrast, we cast coverage planning as conditional sequence generation and directly synthesize temporally ordered trajectory chunks, avoiding post-hoc ordering and stitching.

\paragraph{Sequence-level Generative Models for continuous Trajectory Synthesis}

Sequence-level modeling has been widely studied in imitation learning and structured prediction to generate temporally coherent trajectories over long horizons~\cite{osa2018algorithmic,srinivasan2021fast, duan2024structured}. These approaches emphasize action continuity and executability, addressing the limitations of point-wise or local representations. However, surface coverage is rarely treated as a sequence-level objective in this literature, and continuous generation is typically used to improve motion realism rather than to induce systematic surface traversal. Generative models have recently gained attention for continuous trajectory synthesis in robotics, with diffusion models~\cite{ho2020denoising, janner2022planning} emerging as an effective class due to their iterative refinement of entire trajectories~\cite{chi2023diffusion, carvalho2025motion}. While diffusion-based approaches naturally support global temporal structure and smoothness, existing work primarily targets general control or motion planning tasks~\cite{tevet2022human, saha2024edmp, hoeg2024streaming} rather than coverage-driven surface traversal conditioned on object geometry. Our work aligns generative sequence modeling with the sequence-level requirements of coverage path planning by conditioning diffusion-based trajectory generation on 3D geometry. This enables the direct synthesis of globally ordered trajectories, whose overall structure induces effective surface coverage without relying on heuristic stitching or explicit coverage optimization.

\begin{figure*}
     \centering
    \includegraphics[width=\textwidth]{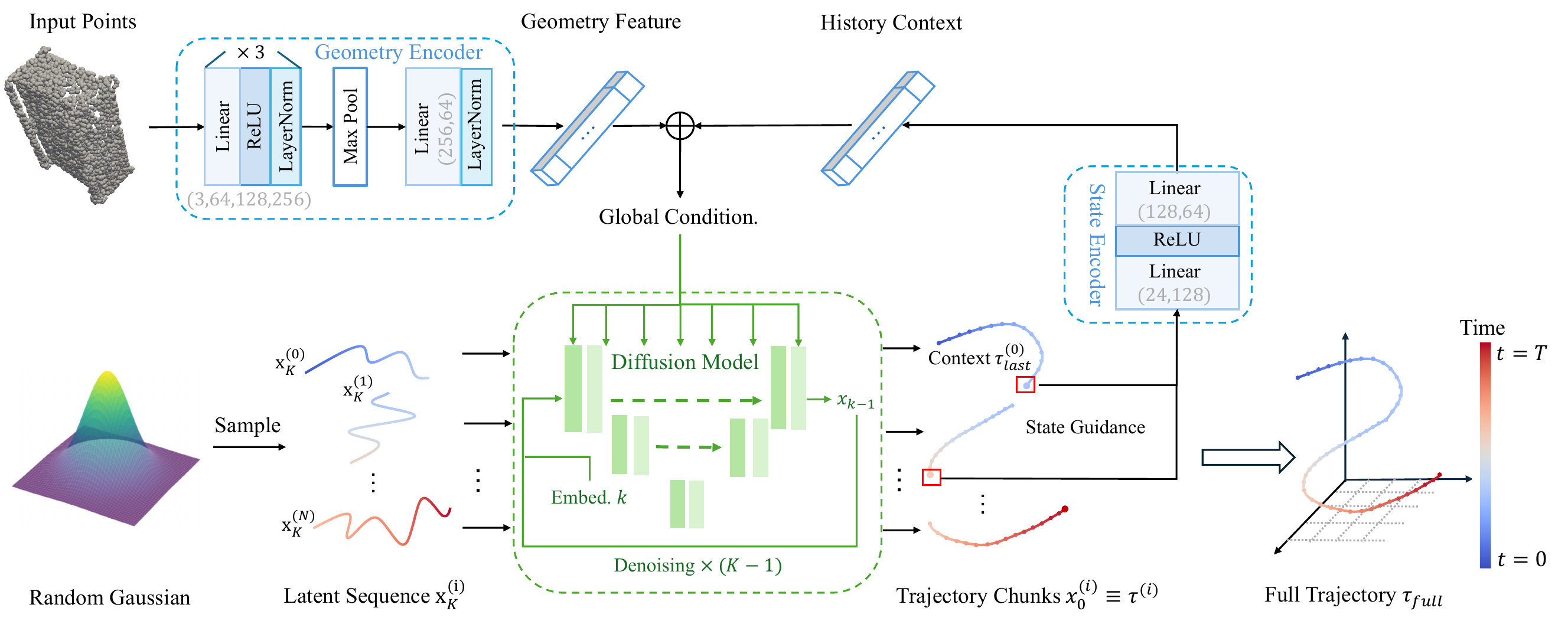}
     \caption{{Illustration of the 3D-CovDiffusion architecture. First, input point clouds are passed through the geometry encoder, which extracts a global observation feature. Simultaneously, the robot state is encoded to produce a state feature. These two features are combined to form the global condition for trajectory generation. Next, a diffusion model samples a noisy trajectory sequence from a Gaussian prior and iteratively denoises it into a noise-free trajectory conditioned on the global features. Finally, the noise-free segments are concatenated to form a complete trajectory.}}
    \label{fig:pipline}
\end{figure*}

\section{METHODOLOGY}
\subsection{Problem Formulation}

We formulate coverage path planning for surface-processing tasks as a sequence-level trajectory generation problem. Given an object observation represented by a 3D point cloud $P$, the goal is to generate a smooth, time-ordered trajectory
\[
\tau = \{a_1, a_2, \dots, a_T\},
\]
where each $a_t \in \mathbb{R}^6$ denotes the 6-DoF end-effector pose at time step $t$. The generated trajectory $\tau$ directly specifies the execution order of surface traversal.

Unlike conventional approaches, we do not impose explicit coverage objectives or segment-wise constraints. Instead, surface coverage is treated as an emergent property of the temporal ordering of the generated trajectory. Under this formulation, coverage quality is determined by how the sequence $\tau$ traverses the surface over time, rather than by local pose accuracy or post-hoc stitching. 


\subsection{Trajectory Representation \& Conditioning}
Following the formulation in Sec.~III-A, the trajectory $\tau$ is represented as an ordered sequence of 6-DoF poses and treated as a single sequence tensor during generation. Rather than predicting unordered local strokes that require complex spatial routing, we represent the motion as a sequence of temporally ordered overlapping chunks, preserving temporal ordering throughout the planning process. Object geometry, represented as a 3D point cloud $P$, is incorporated as a conditioning signal rather than as a prediction target. The geometric condition influences the global structure and ordering of the trajectory, guiding surface traversal at the sequence level instead of applying local corrections to individual poses. Optionally, a short execution history $h=\tau_{1:L_h}$, consisting of the most recent $L_h$ poses, can be provided as additional context to ensure temporal continuity.


\subsection{Sequence-level Generative Framework}
To model coverage path planning at the appropriate abstraction level, we adopt a sequence-level generative modeling framework to learn the conditional distribution $p(\tau \mid P)$. Under this framework, the entire trajectory is generated as a single structured object rather than as a collection of locally predicted actions. Conceptually, trajectory synthesis starts from a random initialization and proceeds through an iterative generation process, producing a complete, ordered sequence in which all time steps are generated jointly.

The sequence-level generative formulation offers several advantages. First, global trajectory ordering is modeled directly rather than recovered through post-hoc sorting or stitching. Second, generating the entire sequence jointly enables long-range surface traversal, which is essential for systematic surface traversal. Finally, this formulation provides a necessary foundation for coverage to emerge as a property of the generated trajectory, rather than as an explicitly optimized objective.


\subsection{Geometry-conditioned Diffusion Policy}
We instantiate the sequence-level generative framework using a geometry-conditioned denoising diffusion implicit model (DDIM)~\cite{song2021denoising}. The diffusion process operates on a trajectory chunk as a sequence and models the conditional distribution $p(\tau^{(m)} \mid P, h_m)$ through iterative refinement (Algorithm~\ref{alg:covdiff}; Fig.~\ref{fig:pipline}), where $\tau^{(m)}$ denotes the $m$-th generated trajectory chunk and $h_m$ denotes its execution-history context. In our setting, each chunk is generated jointly by diffusion, while the full trajectory is obtained by sequentially concatenating chunks. The history $h_m$ is defined as the last $L_h$ poses of the trajectory generated so far, and is used only as a continuity condition for the next chunk.

\paragraph{Forward Process}
The complete trajectory is represented as
\[
\mathbf{x}_0 = \tau^{(m)} \in \mathbb{R}^{H \times 6},
\]
where $H$ is the chunk length and each element is a 6-DoF end-effector pose. In the forward diffusion process, the full trajectory is progressively corrupted with Gaussian noise over a fixed number of diffusion steps:
\[
\mathbf{x}_t
= \sqrt{\bar{\alpha}_t}\,\mathbf{x}_0
+ \sqrt{1-\bar{\alpha}_t}\,\boldsymbol{\epsilon},
\quad
\boldsymbol{\epsilon} \sim \mathcal{N}(\mathbf{0}, \mathbf{I}),
\]
where $t \in \{1,\dots,K\}$ and $\bar{\alpha}_t = \prod_{i=1}^t \alpha_i$ follows a predefined noise schedule. Importantly, this process is applied to the entire trajectory tensor rather than to local segments, preserving the global temporal structure while gradually removing semantic information. We use a predefined noise schedule (e.g., cosine) over $K$ diffusion steps for trajectories of fixed length $T$.

\paragraph{Reverse Denoising Process.}
The reverse process learns to recover a clean trajectory by denoising the entire sequence conditioned on object geometry and execution context, as illustrated in the overall architecture in Fig.~\ref{fig:pipline}. At each diffusion step, a neural denoiser predicts the noise component jointly for all timesteps:
\[
\hat{\boldsymbol{\epsilon}}
= f_\theta(\mathbf{x}_t,\, t,\, c_m),
\quad
c_m = \big[f_{\text{pc}}(P),\; f_{\text{traj}}(h_m)\big],
\]
where $f_{\text{pc}}(P)$ encodes the geometric point cloud and $f_{\text{traj}}(h_m)$ encodes the recent execution history as a continuity context. The condition $c_m$ is injected into the denoiser at every step (e.g., via FiLM modulation), so that geometry and continuity guide global refinement throughout denoising~\cite{perez2018film}. The DDIM reverse update is given by
\[
\mathbf{x}_{t-1}
=
\sqrt{\bar{\alpha}_{t-1}}
\left(
\frac{\mathbf{x}_t - \sqrt{1-\bar{\alpha}_t}\,\hat{\boldsymbol{\epsilon}}}{\sqrt{\bar{\alpha}_t}}
\right)
+
\sqrt{1-\bar{\alpha}_{t-1}}\,\boldsymbol{\eta},
\]
where $\boldsymbol{\eta}$ is an optional noise term (set to zero in DDIM sampling). Generation is sequential only at the chunk level: each chunk is denoised jointly, while the last $L_h$ poses of the accumulated trajectory provide continuity context for the next chunk.

By applying geometry and continuity conditioning at every diffusion step and operating on the full sequence, the geometry-conditioned diffusion policy preserves global temporal coherence, provides implicit smoothness, and avoids error accumulation associated with step-by-step prediction.

\paragraph{Training Objective}
We train $f_\theta$ with the standard DDPM noise-prediction objective, while DDIM is used only for inference-time sampling:
\[
\mathcal{L}_\epsilon = \mathbb{E}_{\mathbf{x}_0,t,\boldsymbol{\epsilon}}
\left[
\left\|\boldsymbol{\epsilon} - f_\theta(\mathbf{x}_t,t,c_m)\right\|_2^2
\right].
\]

\subsection{Implementation Details}
\paragraph{Details of Model Architecture}
We condition the policy on a raw point cloud ($N{=}5120$ points) encoded into a latent geometry embedding (64-d), and use a diffusion model to generate 6-DoF trajectory chunks.
We train with Adam for 200 epochs (batch size 128, learning rate $1\times10^{-4}$), and use DDIM sampling with $K{=}100$ steps and a cosine noise schedule at inference. Unless otherwise stated, we keep the same architecture and hyperparameters across all categories.

To incorporate execution history, we design a trajectory encoder that processes a short historical trajectory segment as a continuity condition. Specifically, a 4-step history of 6-DoF poses (24 dimensions in total) is encoded by a two-layer MLP:
\[
\text{MLP}_{\text{traj}}: \mathbb{R}^{24} \rightarrow \mathbb{R}^{64},
\]
with a Linear(24$\rightarrow$128) layer, ReLU activation, and a Linear(128$\rightarrow$64) layer. The resulting trajectory embedding is concatenated with the geometry embedding to form a 128-dimensional global conditioning vector.

The global condition is injected into the denoising network at every diffusion step using FiLM modulation, which predicts per-channel scale and bias parameters to modulate intermediate features via an affine transformation $\text{FiLM}(x) = \gamma \odot x + \beta$. This conditioning mechanism allows both geometric context and execution history to influence trajectory generation globally rather than through local corrections.

\renewcommand{\algorithmiccomment}[1]{\textsc{#1}}
\begin{algorithm}[t]
\caption{Sequential Geometry-conditioned Diffusion for Coverage Trajectory Generation}
\label{alg:covdiff}
\begin{algorithmic}[1]
\State \textbf{Input:} point cloud $P$, number of chunks $M$, history length $L_h$, DDIM steps $K$
\State \textbf{Output:} generated trajectory $\tau$

\State Encode geometry $z_{\text{pc}} \leftarrow f_{\text{pc}}(P)$
\State Initialize generated trajectory $\tau \leftarrow \emptyset$

\For{$m = 1$ {\bfseries to} $M$}
    \State Set history $h_m \leftarrow$ last $L_h$ poses of $\tau$ \algorithmiccomment{zeros if $m=1$}
    \State Encode continuity context $z_{\text{traj}} \leftarrow f_{\text{traj}}(h_m)$
    \State Form conditioning vector $c_m \leftarrow [z_{\text{pc}}, z_{\text{traj}}]$

    \State Initialize chunk sample $\mathbf{x}^{(m)}_K \sim \mathcal{N}(\mathbf{0}, \mathbf{I})$
    \State Set $\boldsymbol{\eta} \leftarrow \mathbf{0}$ \algorithmiccomment{deterministic DDIM sampling}

    \For{$t = K$ {\bfseries to} $1$}
        \State Predict noise $\hat{\boldsymbol{\epsilon}} \leftarrow f_\theta(\mathbf{x}^{(m)}_t, t, c_m)$
        \State Update chunk using DDIM rule:
        \State \;\; $\mathbf{x}^{(m)}_{t-1} \leftarrow 
        \sqrt{\bar{\alpha}_{t-1}}
        \left(
        \dfrac{\mathbf{x}^{(m)}_t - \sqrt{1-\bar{\alpha}_t}\,\hat{\boldsymbol{\epsilon}}}{\sqrt{\bar{\alpha}_t}}
        \right)
        + \sqrt{1-\bar{\alpha}_{t-1}}\,\boldsymbol{\eta}$
    \EndFor

    \State Append generated chunk $\tau \leftarrow \tau \Vert \mathbf{x}^{(m)}_0$
\EndFor

\State \Return $\tau$
\end{algorithmic}
\end{algorithm}

\paragraph{Training Configuration}
Each training sample consists of a point cloud observation and a short historical trajectory segment, both normalized to object-centric coordinates. The model is trained using the standard noise-prediction loss described in Sec.~III-D. We use the Adam optimizer with a learning rate of $1\times10^{-4}$ and a batch size of 128, and train the model for 200 epochs. No explicit geometric regression or coverage-related loss is introduced during training.

\paragraph{Inference}
At inference, we iteratively generate overlapping trajectory chunks conditioned on object geometry and execution history. Each chunk is produced via joint diffusion denoising, and the full trajectory is obtained by sequential concatenation, avoiding post-hoc heuristic ordering/sorting.

\paragraph{Dataset and Preprocessing}
Experiments are conducted on extended versions of the MaskPlanner datasets~\cite{tiboni2025maskplanner}, which include category-specific expert demonstrations paired with object-centered point clouds for \textit{cuboids}, \textit{windows}, \textit{shelves}, and \textit{containers}. Point clouds and trajectories are scaled and normalized following the baseline preprocessing pipeline. The dataset is split into 80\% for training and 20\% for testing, ensuring that test objects are not observed during training.

\begin{figure*}[htbp]
    \centering
    \includegraphics[width= 1.0 \linewidth]{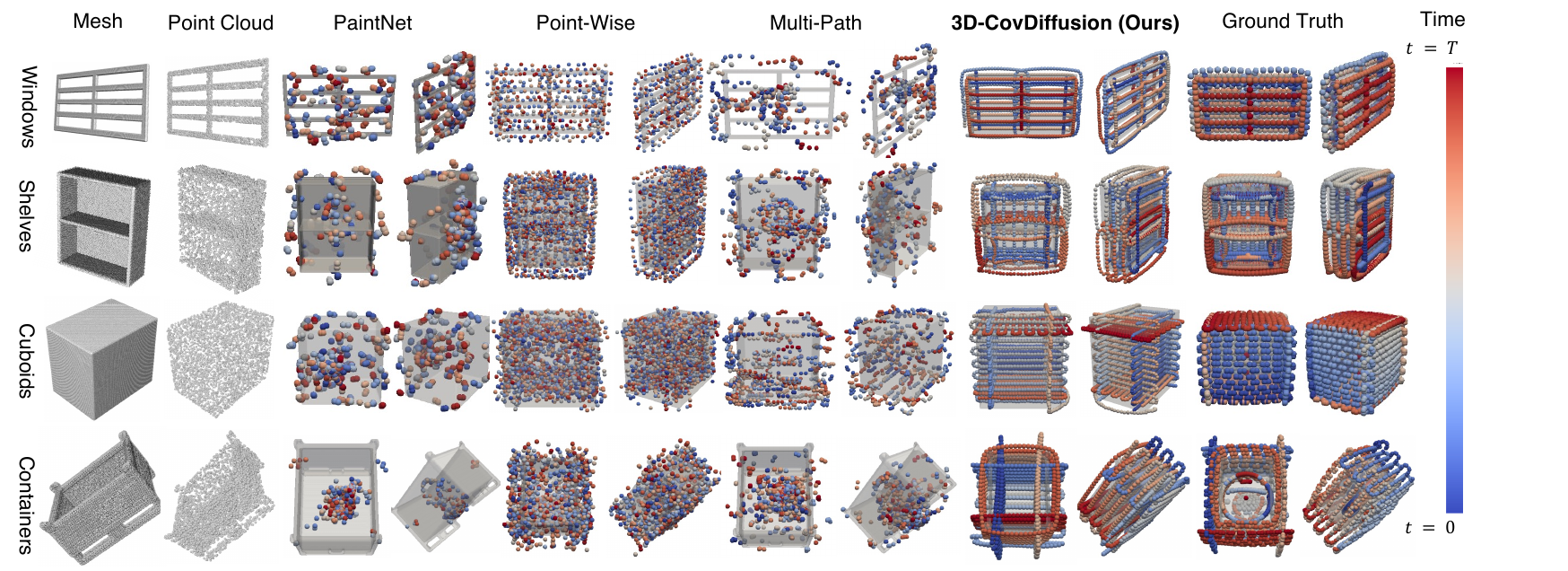}
    \caption{Qualitative comparison of coverage trajectory generation. 
Target objects are shown as dark grey meshes with light grey input point clouds. We compare our 3D-CovDiffusion and the PaintNet baseline against Ground Truth (GT). The blue-to-red gradient indicates normalized execution time $t \in [0, T]$, per the color bar. Containers are evaluated separately as a low-data regime (70 train / 18 test) and are not included in the main-benchmark averages.}
    \label{fig:qualitative_comparison}
\end{figure*}

\begin{figure}
    \centering
    \includegraphics[width=1\linewidth]{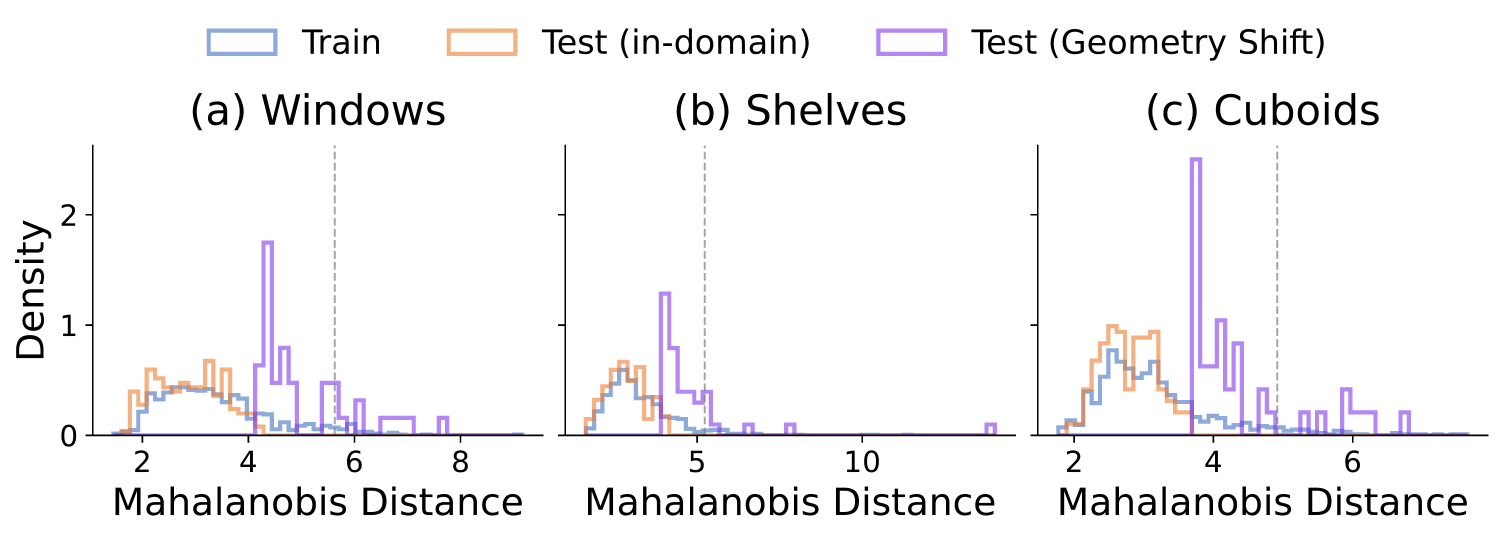}
    \caption{Category-wise Mahalanobis distance distributions for train, in-domain test, and geometry-shifted test on (a) windows, (b) shelves, and (c) cuboids. Each category uses 800/200 train/test; within the 200 test objects, the top 20\% highest-distance instances are marked as geometry-shifted (40 OOD) and the remaining 160 as in-domain. The dashed line indicates the 95th percentile of training distances (sensitivity analysis).}
    \label{fig:setup}
    \vspace{-2em}
\end{figure}

\section{EXPERIMENTAL RESULTS}
\subsection{Experimental Setup}
We evaluate on industrial robotic spray painting using expert demonstrations paired with object-centered 3D point clouds. Each sample contains a point-cloud observation and an expert 6-DoF end-effector trajectory targeting high surface coverage with smooth, time-ordered motions. The dataset includes four categories: \textit{windows}, \textit{cuboids}, \textit{shelves}, and \textit{containers}.

We use category-wise 80/20 train/test splits with disjoint objects: \textit{windows}/\textit{cuboids}/\textit{shelves} each have 1000 objects (800/200), while \textit{containers} has 88 (70/18) and is reported separately as a low-data evaluation to avoid high-variance results confounding the main benchmark.

Unless otherwise specified, a single model is trained jointly across all categories and evaluated without category-specific fine-tuning (category-agnostic zero-shot).
For windows, cuboids, and shelves, we define a distribution-shifted test subset (geometry shift) within each category using Mahalanobis distance in a geometric feature space computed from training statistics: among the 200 test objects, we select the top 20\% highest-distance instances as geometry-shifted (40 OOD), and treat the remaining 160 as in-domain test (see Fig.~\ref{fig:setup}).
As a sensitivity check, we observe consistent trends when alternatively thresholding by the 95th percentile of training distances. All baselines~\cite{tiboni2025maskplanner,tiboni2023paintnet} follow the same splits and evaluation protocols, so differences reflect modeling choices rather than experimental conditions.

\subsection{Baselines and Evaluation Metrics}
We compare the proposed approach against representative baseline methods that reflect different modeling assumptions for surface-processing trajectory generation. All baselines are evaluated using the same data splits, input representations, and evaluation protocols to ensure a fair comparison.

\paragraph{Segment-wise Learning-based Methods}
This class includes approaches that predict short trajectory strokes independently and assemble them using heuristic ordering or stitching. While effective for modeling local motion patterns, these methods do not explicitly model global temporal ordering, making surface coverage a post-hoc outcome rather than an intrinsic property of the generated trajectory.

\paragraph{Point-wise Trajectory Regression}
Point-wise baselines predict individual poses conditioned on local observations. Although they can achieve low geometric error at the pose level, they lack mechanisms for modeling long-range temporal dependencies and therefore cannot guarantee systematic surface traversal.

\paragraph{Multi-path Variants}
We additionally consider variants that generate multiple candidate paths and select among them using heuristic criteria. These baselines help isolate whether coverage limitations stem from insufficient path diversity or from the underlying segment-wise modeling assumption.

\begin{table*}[t]
\centering
\caption{
Quantitative results across three metrics: Point-wise Chamfer Distance (PCD), 
Surface Coverage Rate (Coverage, reported in \%), and Smoothness. 
Values are reported as mean $\pm$ standard deviation over three random seeds 
(standard deviations smaller than $0.005$ are omitted). 
}
\label{tab:main_quantitative_results}

\setlength{\tabcolsep}{3.5pt}
\renewcommand{\arraystretch}{1.15}
\small

\resizebox{\textwidth}{!}{%
\begin{tabular}{l|ccc|ccc|ccc}
\toprule
\multicolumn{1}{l|}{\textbf{Category}} 
& \multicolumn{3}{c|}{\textbf{Windows}} 
& \multicolumn{3}{c|}{\textbf{Cuboids}} 
& \multicolumn{3}{c}{\textbf{Shelves}} \\
\cmidrule(lr){2-10} 
\textbf{Model}
& PCD & Coverage & Smoothness
& PCD & Coverage & Smoothness
& PCD & Coverage & Smoothness \\
\midrule
Point-Wise
& $55.71 \pm 3.10$ & $96.49 \pm 0.11$\% & $2.54$
& $35.47 \pm 0.22$ & $98.73 \pm 0.11$\% & $3.50$
& $44.30 \pm 0.59$ & $92.96 \pm 0.02$\% & $2.20$ \\

Multi-Path
& $264.69 \pm 0.43$ & $67.54 \pm 0.13$\% & $1.69$
& $297.40 \pm 0.27$ & $50.66 \pm 0.15$\% & $3.30$
& $491.86 \pm 3.29$ & $26.05 \pm 0.32$\% & $0.72$\\

PaintNet
& $694.88 \pm 76.17$ & $63.57 \pm 0.16$\% & $1.37$
& $689.84 \pm 1.71$ & $11.47 \pm 0.14$\% & $1.32$
& $744.64 \pm 3.72$ & $16.22 \pm 0.80$\% & $0.26$ \\

\textbf{3D-CovDiffusion (Ours)}
& $\mathbf{10.24} \pm 0.49$ & $\mathbf{99.91} \pm 0.05$\% & $\mathbf{0.05}$
& $\mathbf{4.82} \pm 0.14$ & $\mathbf{98.88} \pm 0.08$\% & $\mathbf{0.04}$
& $\mathbf{9.01} \pm 0.30$ & $\mathbf{95.03} \pm 0.30$\% & $\mathbf{0.07}$ \\

\bottomrule
\end{tabular}%
}
\end{table*}

\paragraph{Evaluation Metrics}
We evaluate all methods using three complementary metrics. Point-wise Chamfer Distance (PCD) is used to assess geometric fidelity between predicted and reference trajectories at the pose level~\cite{pcd}. Given two point sets $S_1, S_2 \subseteq \mathbb{R}^3$, the symmetric Chamfer Distance is defined as:
\begin{equation}
d_\text{CD}(S_1, S_2) = \sum_{x \in S_1} \min_{y \in S_2} \|x - y\|_2^2 + \sum_{y \in S_2} \min_{x \in S_1} \|x - y\|_2^2.
\end{equation} 
All point clouds and trajectories are normalized to object-centric coordinates, and PCD is computed in the normalized space for all methods.
\begin{figure}[t]
    \centering
    \includegraphics[width= 1 \linewidth]{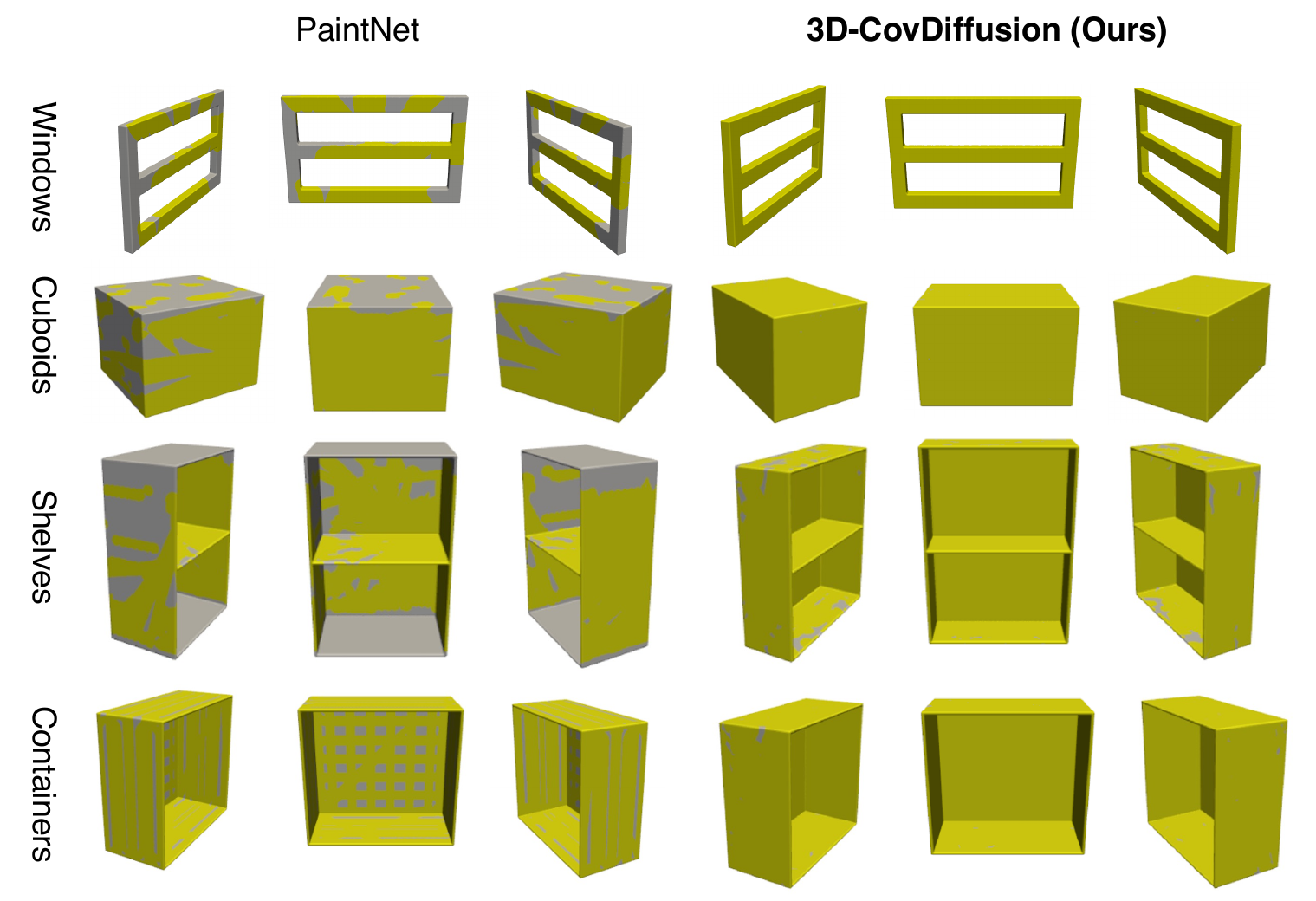}
    \caption{Qualitative coverage comparison across object categories. Columns (left to right) show PaintNet, and 3D-CovDiffusion (Ours); rows correspond to Windows, Cuboids and Shelves. Each cell presents multiple representative viewpoints with surface coverage visualization: yellow regions indicate covered/painted surfaces, while gray regions indicate uncovered surfaces. This facilitates visual comparison of coverage completeness and consistency across methods.}
    \label{fig:coverage}
    \vspace{-2em} 
\end{figure}

Surface Coverage Rate is used as the primary evaluation metric, as it directly reflects task-relevant completeness in surface-processing applications. A surface face $f_j$ is considered covered if the minimum distance between its centroid $\mathbf{c}_j$ and any trajectory segment $(\mathbf{p}_s, \mathbf{p}_e)$ is within a predefined spray radius $r_{\text{spray}}$:
\begin{equation}
d(\mathbf{c}_j, \mathbf{S}) = \min_{(\mathbf{p}_s,\mathbf{p}_e) \in \mathbf{S}}
\left\| \mathbf{c}_j - \Pi_{(\mathbf{p}_s,\mathbf{p}_e)}(\mathbf{c}_j) \right\|_2 \le r_{\text{spray}},
\end{equation}
where $\Pi_{(\mathbf{p}_s,\mathbf{p}_e)}(\cdot)$ denotes projection onto the line segment. The overlapping coverage rate is computed as:
\begin{equation}
C^{\text{overlap}} =
\frac{1}{|F|}
\sum_{f_j \in F}
\mathbb{1}\!\left[d(\mathbf{c}_j, \mathbf{S}) \le r_{\text{spray}}\right].
\end{equation}

We additionally report area-weighted coverage, defined by weighting each face by its area:
\begin{equation}
C^{\text{area}} =
\frac{\sum_{f_j \in F} A(f_j)\,\mathbb{1}\!\left[d(\mathbf{c}_j,\mathbf{S}) \le r_{\text{spray}}\right]}
{\sum_{f_j \in F} A(f_j)}.
\end{equation}
where $A(f_j)$ denotes the area of face $f_j$. Trajectory smoothness is measured using translational jerk. For the position sequence $\{\mathbf{p}_t\}_{t=1}^{T}$ of a generated 6-DoF trajectory, we define
\begin{equation}
J =
\frac{1}{T-3}\sum_{t=2}^{T-2}
\left\|
\mathbf{p}_{t+2}-3\mathbf{p}_{t+1}+3\mathbf{p}_{t}-\mathbf{p}_{t-1}
\right\|_2^2 .
\end{equation}
Lower $J$ indicates smoother motion. Jerk is computed on the translational components in normalized object-centric coordinates; orientations are not included in this metric. Coverage is treated as the primary metric, while PCD and jerk assess geometric accuracy and motion continuity. All metrics are computed with the same protocol across methods.

\begin{figure}[t]
    \centering
    \includegraphics[width=1\linewidth]{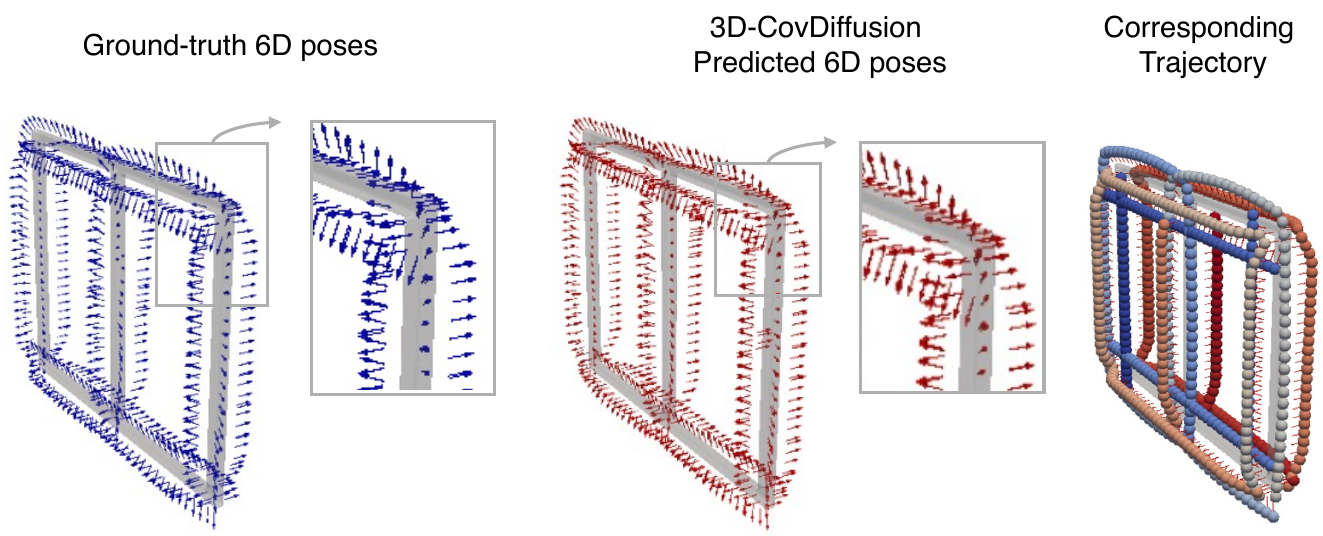}
    \caption{Predicted 6D poses (red) through our method compared with Ground Truth 6D poses (blue). Pose orientations are efficiently preserved and learned.}
    \label{fig: 6D-orientation}
    \vspace{-1em} 
\end{figure}
\begin{figure}[ht]
    \centering
    \includegraphics[width=1.0\linewidth]{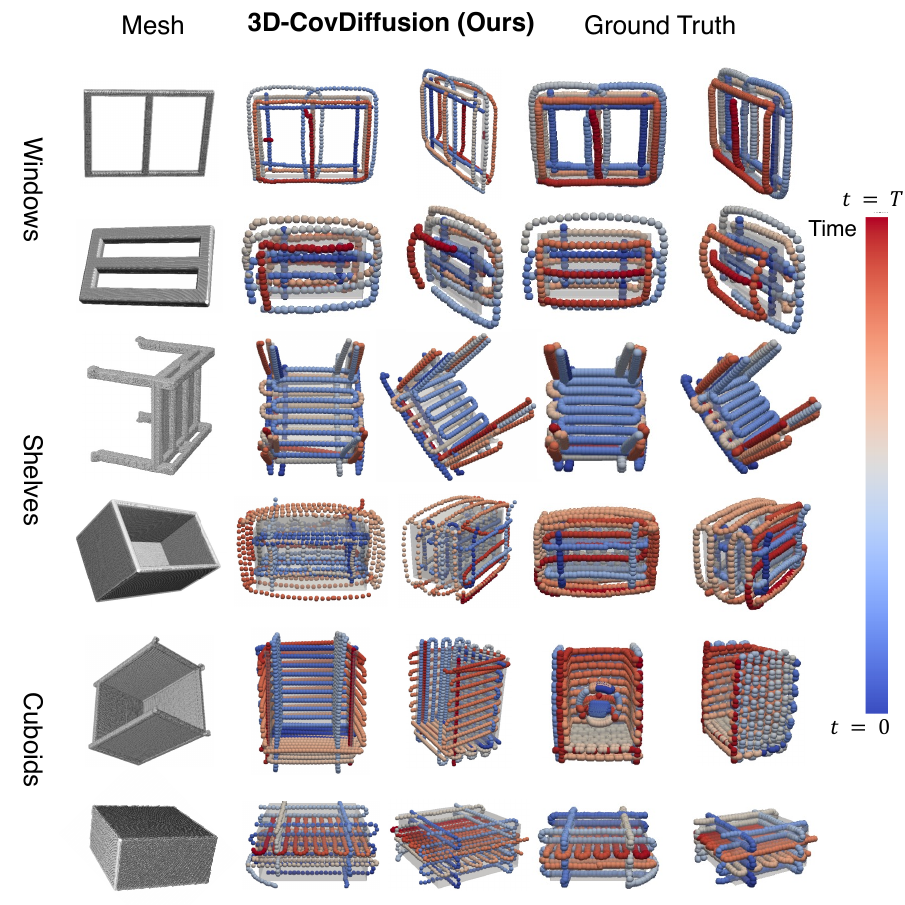}
    \caption{Distribution-shifted qualitative comparison (geometry shift) for coverage trajectory generation. Columns (left→right): mesh, 3D-CovDiffusion (ours), Ground Truth. Trajectories are colored by normalized time $t\in[0,T]$ (blue→red). }
    \label{fig:geometry_shift}

\end{figure}
\subsection{Main Results: Coverage-aware Trajectory Generation}


\paragraph{Main Qualitative Results}
Fig.~\ref{fig:qualitative_comparison} qualitatively compares generated trajectories across the benchmark categories. PaintNet and Multi-Path frequently produce fragmented patterns with weak temporal structure, while 3D-CovDiffusion yields temporally coherent, time-ordered paths that visually match the systematic traversal in the expert ground truth. Point-Wise, in contrast, often achieves dense surface alignment but appears as an unordered set of poses without coherent traversal order, making smooth execution difficult.

To further inspect motion precision, Fig.~\ref{fig: 6D-orientation} shows a close-up comparison of predicted and ground-truth 6-DoF poses. The visualization indicates that our model captures both positions and orientations, keeping the spray tool consistently aligned with local surface geometry. The coverage visualization in Fig.~\ref{fig:coverage} further highlights the systematic nature of our approach: sequence-level generation produces continuous sweep patterns with fewer gaps compared to segment-/path-based baselines that rely on post-hoc ordering/stitching. Fig.~\ref{fig:geometry_shift} shows that this behavior persists under within-category geometry shift, where baselines often degrade into fragmented trajectories on shifted geometries, while 3D-CovDiffusion maintains global coherence coverage.

\begin{table}[ht]
    \centering
    \caption{Low-data evaluation on Containers (70/18 training instances). Metrics: PCD, Coverage (\%), and Smoothness. Values are mean $\pm$ std over three seeds (std $<0.005$ omitted).}
    \label{tab:containers_lowdata}
    \renewcommand{\arraystretch}{1.0}
    \setlength{\tabcolsep}{4pt}
    \resizebox{0.95\linewidth}{!}{%
    \begin{tabular}{lccc}
    \toprule
    \textbf{Model} & \textbf{PCD} & \textbf{Coverage} & \textbf{Smoothness} \\
    \midrule
    Point-Wise & $\mathbf{313.98} \pm 2.47$ & $\mathbf{82.86}\pm 0.60$\% & $1.71$ \\
    Multi-Path & $1193.45 \pm 5.97$ & $14.12 \pm 0.80$\% & $0.33$ \\
    PaintNet & $2621.26 \pm 34.64$ & $1.54 \pm 0.25$\% & $0.17$ \\
    \textbf{3D-CovDiffusion (Ours)} & $622.16 \pm 26.62$ & $71.10 \pm 1.22$\% & $\mathbf{0.04}$ \\
    \bottomrule
    \end{tabular}%
    }
    \vspace{-1em}
\end{table}

\paragraph{Main Quantitative Results}
Table~\ref{tab:main_quantitative_results} summarizes the performance across categories in terms of surface coverage, point-wise Chamfer Distance (PCD), and trajectory smoothness. In these tasks, coverage and PCD jointly define success by measuring systematic traversal and geometric alignment, respectively, while smoothness reflects motion continuity.



Across the main benchmark categories, prior learning-based baselines reveal a tension between geometric fidelity, coverage, and motion continuity. Point-wise regression attains high coverage but exhibits substantially higher jerk (worse smoothness). In contrast, segment-/path-based baselines such as PaintNet and Multi-Path rely on post-hoc ordering/stitching yet still underperform markedly in coverage and geometric fidelity (high PCD). By generating temporally ordered trajectory chunks at the sequence level, our method improves overlapping coverage, PCD-based geometric alignment, and smoothness simultaneously, without introducing an explicit coverage objective or heuristic post-processing. This trend is consistent across categories with different geometry profiles (e.g., cuboids and shelves), suggesting category-agnostic behavior within the benchmark. Since \textit{Containers} is a low-data regime (70 train / 18 test), we report it separately (Table~\ref{tab:containers_lowdata}) to avoid high-variance results confounding the main benchmark.



\subsection{Ablation Studies}
\paragraph{Effect of Geometry Encoder Backbone}
To assess the influence of point-cloud encoder backbones on downstream trajectory generation, we performed an ablation on the \textit{Windows} category comparing four encoders: PointNet \cite{qi2017pointnet}, PointNet++ \cite{qi2017pointnet++}, Point Transformer \cite{zhao2021point}, and our proposed 3D-CovDiff Encoder. All encoders were trained under the same protocol as the main experiments: training for 200 epochs. Evaluation metrics include PCD, two coverage measures (overlapping coverage and area-weighted coverage), and Smoothness. The numeric results are presented in Table~\ref{tab:encoder-ablation}, which reports an ablation study on different point cloud encoder backbones for the windows category. While stronger encoders such as PointNet++ and Point Transformer improve geometric representation compared to a basic PointNet, the overall differences among encoder choices are relatively modest. Importantly, the proposed framework consistently achieves high coverage (higher is better) and low PCD (lower is better) across all encoder backbones, indicating that performance gains are not primarily driven by encoder complexity. This suggests that sequence-level trajectory generation and coverage-aware diffusion is more important than the specific choice of point cloud encoder.
\begin{table}[ht]
    \centering
    \caption{Ablation study on point cloud encoder backbones for the Windows category.}
    \label{tab:encoder-ablation}
    \renewcommand{\arraystretch}{1.0}
    \setlength{\tabcolsep}{4pt}
    \resizebox{0.95\linewidth}{!}{%
    \begin{tabular}{lcccc}
    \toprule
    \multirow{2}{*}{Encoder Backbone} & \multirow{2}{*}{PCD} & \multicolumn{2}{c}{Coverage} & \multirow{2}{*}{Smoothness} \\
    & & Overlapping & Area-weighted & \\
    \midrule
    PointNet        & 38.55 & \textbf{77.49\%} & 98.83\% & 0.0505 \\
    PointNet++      & 31.77 & 76.09\% & 99.23\% & 0.0522 \\
    Point Transformer & 25.25 & 76.84\% & 98.71\% & 0.0490 \\
    \textbf{3D-CovDiffusion Encoder (Ours)} & \textbf{10.41} & 75.95\% & \textbf{99.84\%} & \textbf{0.0391} \\
    \bottomrule
    \end{tabular}%
    }
\end{table}

\paragraph{Effect of Execution History Conditioning}
To evaluate the contribution of trajectory-aware conditioning in our diffusion model, we conduct an ablation on the \textit{Windows} category by comparing four variants: (1) Previous Traj. (Ours) full model conditioned on the last-point trajectory; (2) Zero Traj. The trajectory input is replaced by an all-zero vector of the same dimensionality; (3) No Traj, trajectory encoder removed. All variants were trained for 200 epochs, and results are shown in Table~\ref{tab:traj-cond-ablation}, which evaluates that removing trajectory conditioning or using zero trajectory input leads to substantial degradation in coverage, geometric alignment, and trajectory smoothness. In contrast, conditioning on the executed trajectory history significantly improves all metrics. This confirms that execution history provides essential continuity context for sequence-level diffusion, stabilizing temporal ordering and enabling systematic surface traversal rather than decomposing the task into independent segments.
\begin{table}[ht]
    \centering
    \caption{Ablation study on trajectory-conditioned diffusion model variants for the Windows category.}
    \label{tab:traj-cond-ablation}
    \renewcommand{\arraystretch}{1.1}
    \setlength{\tabcolsep}{3pt}
    \begin{tabular}{lcccc}
    \toprule
    \multirow{2}{*}{Trajectory Input} & \multirow{2}{*}{PCD} & \multicolumn{2}{c}{Coverage} & \multirow{2}{*}{Smoothness} \\
    & & Overlapping & Area-weighted & \\
    \midrule
    Zero Traj. & 284.46 & 64.10\% & 90.77\% & 0.1894 \\
    No Traj.(remove) & 246.07 & 71.82\% & 90.80\% & 0.1757 \\
    \textbf{Previous Traj. (Ours)} & \textbf{10.41} & \textbf{75.95\%} & \textbf{99.84\%} & \textbf{0.0391} \\
    \bottomrule
    \end{tabular}
\end{table}

\subsection{Discussion and Failure Cases}
The experimental results suggest that sequence-level, geometry-conditioned diffusion can generate trajectories for coverage path planning across the benchmark categories, but limitations appear under challenging conditions. We observe failures on objects with highly complex geometry or sharp curvature, where noisy point clouds provide incomplete conditioning and can cause local deviations from the surface.

\section{CONCLUSIONS}
This paper presents a sequence-level, geometry-conditioned diffusion framework for trajectory generation in industrial surface-processing tasks. We cast coverage path planning as conditional sequence generation and synthesize temporally ordered trajectory chunks directly from raw 3D point clouds, avoiding post-hoc heuristic ordering and stitching used in prior learning-based methods. On the benchmark categories, the proposed approach improves overlapping coverage, geometric alignment, and trajectory smoothness over prior baselines. These findings underscore the importance of modeling trajectory ordering at the sequence level for coverage-driven tasks. Future work will explore richer geometric observations and adaptive history conditioning to improve robustness under noisy inputs.





\end{document}